\documentclass[10pt,twocolumn,letterpaper]{article}

\usepackage{iccv}
\usepackage{times}
\usepackage{epsfig}
\usepackage{graphicx}
\usepackage{amsmath}
\usepackage{amssymb}

\usepackage[accsupp]{axessibility}
\usepackage{graphicx}
\usepackage{amsmath}
\usepackage{amssymb}
\usepackage{booktabs}
\usepackage{multirow}
\usepackage{makecell}
\usepackage{gensymb}
\usepackage{bm}
\usepackage[ruled]{algorithm}
\usepackage{algpseudocode}
\usepackage{lipsum}
\usepackage{soul}
\usepackage{color}
\usepackage{colortbl}
\usepackage[table,xcdraw]{xcolor}

\newcommand{\ETH}{$\mathcal{D}_E$}
\newcommand{\GazeC}{$\mathcal{D}_G$}
\newcommand{\MPII}{$\mathcal{D}_M$}
\newcommand{\ED}{$\mathcal{D}_D$}

\newcommand{\proposed}{PCFGaze}
\newcommand{\prp}{PCFGaze}
\newcommand{\feature}{Physics-Consistent Feature}
\newcommand{\ft}{PCF}
\newcommand{\proposeda}{Spherical Fitting}
\newcommand{\pa}{SF}
\newcommand{\proposedb}{PCF-Oriented Training}
\newcommand{\pb}{IT}
\newtheorem{hyp}{Hypothesis}

\newcommand{\upscore}[1]{\footnotesize{$\blacktriangledown$ #1\%}}
\newcommand{\upscorered}[1]{\footnotesize{$\blacktriangledown$ \underline{#1\%}}}

\newcommand{\marka}[1]{\textbf{#1}}

\newcommand{\std}[1]{\footnotesize{$\pm #1$}}

\usepackage[pagebackref=true,breaklinks=true,letterpaper=true,colorlinks,bookmarks=false]{hyperref}
\usepackage[capitalize]{cleveref}
\crefname{section}{Sec.}{Secs.}
\Crefname{section}{Section}{Sections}
\Crefname{table}{Table}{Tables}
\crefname{table}{Tab.}{Tabs.}
\usepackage[breaklinks=true,bookmarks=false]{hyperref}

\iccvfinalcopy % *** Uncomment this line for the final submission

\def\iccvPaperID{****} % *** Enter the ICCV Paper ID here

\ificcvfinal\fi

\begin{document}

%%%%%%%%% TITLE
\title{PCFGaze: Physics-Consistent Feature for Appearance-based Gaze Estimation}

\author{Yiwei Bao, Feng Lu\textsuperscript{\rm} 
\thanks{Corresponding Author.}\\
State Key Laboratory of VR Technology and Systems, School of CSE, Beihang University\\
{\tt\small \{baoyiwei, fenglu\}@buaa.edu.cn}
% \and
% Feng Lu\\
% State Key Laboratory of VR Technology and Systems, School of CSE, Beihang University\\
% {\tt\small lufeng@buaa.edu.cn}
}
% \author{Yiwei~Bao\textsuperscript{\rm 1}quad~Feng~Lu\textsuperscript{\rm 1, 2} 
% % \thanks{Corresponding Author.  This work was supported by the National Natural Science Foundation of China (NSFC) under Grant 61972012.}\\
% {\textsuperscript{\rm 1} State Key Laboratory of VR Technology and Systems, 
%   School of CSE, Beihang University}  \\
%  {\textsuperscript{\rm 2} Peng Cheng Laboratory, Shenzhen, China}\\
% {\tt\small \{baoyiwei, lufeng\}@buaa.edu.cn\quad~wanghf@pcl.ac.cn}

\maketitle
\ificcvfinal\thispagestyle{empty}\fi

%%%%%%%%% ABSTRACT
\begin{abstract}
   Although recent deep learning based gaze estimation approaches have achieved much improvement, we still know little about how gaze features are connected to the physics of gaze. In this paper, we try to answer this question by analyzing the gaze feature manifold. Our analysis revealed the insight that the geodesic distance between gaze features is consistent with the gaze differences between samples. According to this finding, we construct the Physics-Consistent Feature (PCF) in an analytical way, which connects gaze feature to the physical definition of gaze. We further propose the PCFGaze framework that directly optimizes gaze feature space by the guidance of PCF. Experimental results demonstrate that the proposed framework alleviates the overfitting problem and significantly improves cross-domain gaze estimation accuracy without extra training data. The insight of gaze feature has the potential to benefit other regression tasks with physical meanings.
\end{abstract}

%%%%%%%%% BODY TEXT
\section{Introduction}
\label{sec:intro}

Eye gaze reveals where human attention lands, which has been widely applied in a variety of territories, such as VR/AR systems \cite{burova2020utilizing, konrad2020gaze, sitzmann2018saliency}, medical analysis \cite{castner2020deep, king2020leveraging, kerr2019eye}~and human-computer interaction\cite{kyto2018pinpointing, WangYWWL20, stellmach2011}. Gaze estimation methods can be categorized into two types, model-based approaches and appearance-based approaches. Both approaches have its own pros and cons. Model-based approaches estimate gaze by modeling the anatomical structure of eyeball, which have achieved remarkable accuracy as they are in line with physical rules behind the gaze. Unfortunately, they usually require dedicated hardware such as infrared cameras and light sources. Appearance-based approaches use inexpensive web cameras, where they employ Convolutional Neural Networks (CNNs) to regress gaze direction from user face/eye images. Although these work achieve satisfactory performance within the same dataset, their accuracy severely degrades in cross-domain settings.
\begin{figure}
\begin{center}
% \begin{overpic} 
% [width=\linewidth]
% {example-image-a}
% \end{overpic}
\includegraphics[width=\linewidth]{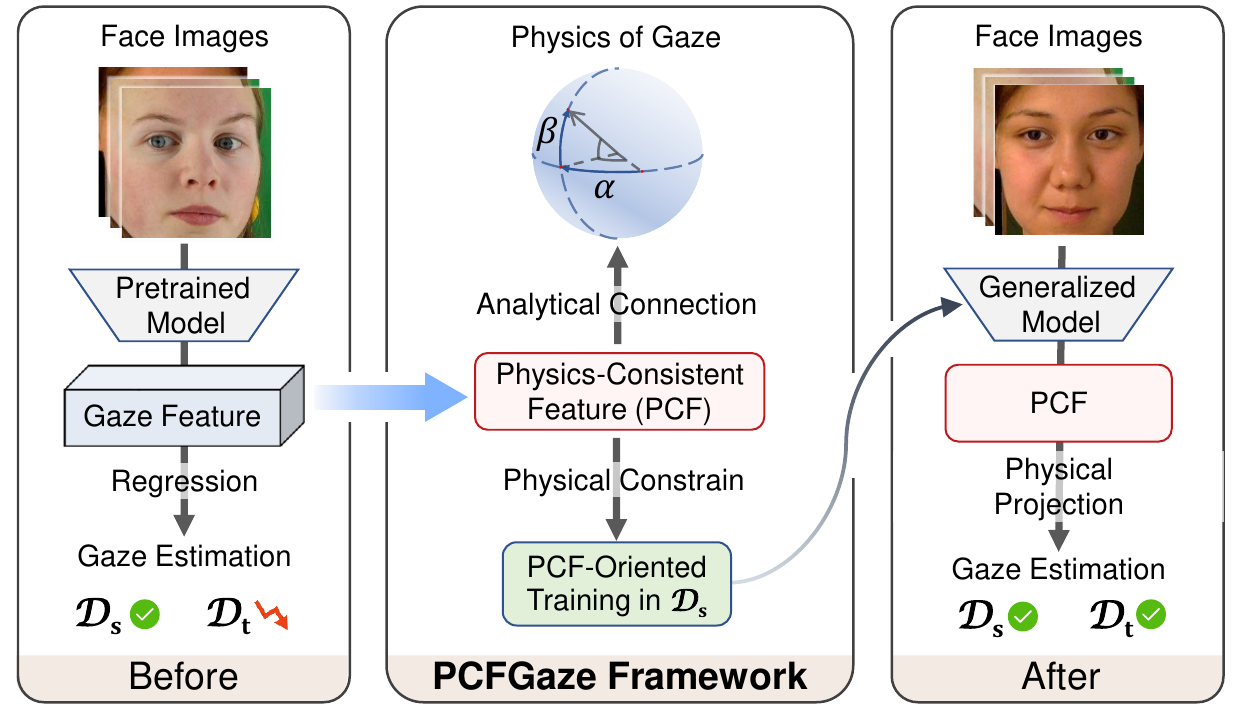}
\end{center}
% \vspace{-2mm}
\caption{
We propose the Physics-Consistent Feature for appearance-based gaze estimation (\prp). We connect gaze feature to the physics of gaze by constructing the \feature~(\ft). Then, we retrain the model in source domain by the constrain of physics provided by \ft. The generalized model estimate gaze in an analytical way and improves the generalization ability of gaze estimation models.
}
\label{fig:Teaser}
% \vspace{-2mm}
\vspace{2mm}
\end{figure}

To improve the cross-domain performance, various domain adaptation approaches have been proposed, \ie, adversarial learning \cite{kellnhofer2019gaze360,wang2019generalizing}, contrastive learning \cite{wang2022contrastive}~and collaborative learning \cite{liu2021generalizing}. However, these approaches require a number of target domain samples for adaptation, which is limited in real-world applications, as it is not feasible for users to collect target domain data and re-optimize the network.

On the contrary, gaze generalization approaches are user-friendly since it requires no target domain data. For example, Cheng \etal~proposed to generalize gaze estimation model by purifying gaze feature during source domain training \cite{cheng2022puregaze}. In fact, gaze generalization task is more challenging since the constraints from target domain is missing.

To bridge the gap between the source/target domain when the target domain data is inaccessible, we argue that the underlying principles behind the gaze in two domains should be discovered. In other words, we have to find some principles to constrain the model as an alternative of target domain data. However,
such issue has been less investigated in previous studies.
% The core question to a generalizable gaze estimation model is that, what principles should we follow to constrain the model? %The main difficulty is how to find a better constrain than gaze label for generalizable model training.

In this paper, we aim to address the above challenge by a simple-but-effective idea, \ie, to optimize the model based on the physical rules of gaze instead of target domain data. While previous approaches are all data-driven, our idea is essentially to break the data-driven pipeline, but to introduce the physical-rule-driven approach. We believe that with the proper criteria, the extracted gaze feature is capable of revealing consistent numerical patterns of gaze. This connection helps us to constrain gaze feature by the physical rules of gaze, which alleviates the overfitting problem by reducing the dependence on ground truth data.  
% In this paper, we aim to answer the above question by analyzing the relation between gaze feature and gaze rotations. Based on the observation that gaze is an rotation with two degrees of freedom, we propose the hypothesis that high dimensional gaze feature manifold could be mapped to the surface of a sphere in 3D space. Our experiments shown that the geodesic distance between feature is proportional to the gaze differences between samples. Thus, we construct the proposed \feature~(\ft) by Isometric map algorithm. The proposed \ft~distributes on the surface of a sphere and share the same gaze distribution pattern as the physical definition of gaze. \ft~connect the feature with the physics of gaze by feature' intrinsic property, which may optimize feature space by the physics of gaze and alleviate the overfitting problem.

Following the above idea, we propose the \proposed~framework for generalizable gaze estimation. First, we construct the \feature~(\ft)~based on feature obtained from pretrained model in an unsupervised manner. The proposed \ft~shares the same gaze distribution pattern with the physical definitions of gaze. Second, the \ft~optimizes gaze feature with source domain labels based on the physical rules of gaze to address the overfitting issue. The primary contributions of this work are as follow:
% In light of the above findings, we propose the \proposed~(\prp) framework for generalizable gaze estimation. The proposed \prp~framework optimize the feature space through the help of \ft. The \prp~framework consists of two phases: the \proposeda~(\pa) phase and the \proposedb~(\pb) phase. In the \pa~phase, we model the gaze distribution of \ft~according to the physics of gaze so that gaze is estimated from feature in an analytical way. In the \pb~phase, we integrate Isomap to the model training process by introducing the proposed Isometric Propagator, so that the gaze feature space is optimized with the guidance of \ft. 
% Experiments shown that \inred{TODO}.

\begin{itemize}
    \item We propose the \feature~ that connects gaze feature to the physical definition of gaze without label. The insight brought by \ft~may also inspire other regression tasks with physical meaning.
    \item We propose the \prp~framework for generalizable gaze estimation. It estimates gaze analytically and optimizes the feature space via the physical rules of gaze.
    \item Experimental results illustrate that the proposed framework achieves consistent improvements in 7 different cross-dataset settings. The \prp~improves the generalization ability of baseline model up to $29.23\%$ without touching target domain data.
\end{itemize}
%------------------------------------------------------------------------
\section{Related Work}
\subsection{Gaze Estimation}
%  \subsubsection{Model-based Methods}
There are two mainstream gaze estimation approaches, the model-based approach and the appearance-based approach.
%Model-based approaches estimate gaze analytically. By detecting the position of pupil and iris, these methods model the anatomy structure of eye and calculate the rotation angle of eyeballs \cite{hansen2009eye}.
Model-based approaches estimate gaze by reconstructing the anatomy structure of the eyeball \cite{hansen2009eye}.
These methods achieve remarkable accuracy but also require personal calibration and dedicated devices such as 
 depth sensors \cite{sun2015real, xiong2014eye}, infrared cameras \cite{takemura2017hybrid, guestrin2006general}~and lights \cite{guestrin2006general, liu2022method}~to accurately rebuild the tiny structures like pupil and corneal.

% \subsubsection{Appearance-based Methods}
Appearance-based approaches estimate gaze from user images captured by a single web camera. Early methods estimate gaze from eye images by traditional machine learning algorithms like manifold embedding \cite{schneider2014manifold}~and adaptive linear regression \cite{lu2014adaptive}. Lu \etal~propose to estimate eye rotation by measuring the geodesic distance between eye images \cite{lu2017appearance}. More recently, a number of gaze estimation datasets have been collected \cite{funes2014eyediap, krafka2016eye, zhang2017s, kellnhofer2019gaze360, zhang2020eth}. These datasets provide hundreds of thousands of user images with gaze labels, which makes deep learning based gaze estimation possible. Representative studies include gaze estimation using convolutional neural networks (CNNs) \cite{zhang2015appearance} with eye images \cite{zhang2015appearance, cheng2020gaze} or face images \cite{zhang2017s, cheng2020coarse, chen2018appearance, krafka2016eye, bao2021adaptive}. 
Some previous studies also represent gaze features as low dimensional manifolds for personalization \cite{park2019few}~and unsupervised learning \cite{yu2020unsupervised}. But these methods still construct manifolds by data-driven learning approach with supervision like gaze redirection. Our method connects gaze feature to 3D space by the physical rules in an unsupervised manner.
%These deep-learning based method achieves 
% By training with large amount of ground truth data, deep-learning based methods achieves satisfactory results within dataset. However, these methods can easily overfit to gaze-irrelevant features such as illumination and human appearance, leading to severe performance degradation when testing on a different dataset.

\subsection{Cross-domain Gaze Estimation}
One of the major problem of the deep-learning based approaches is that the performance degrade severely when testing on a different domain.
%The performance degradation of the deep learning based approaches has drawn much research attention in recent years. 
To improve the cross-domain performance, a number of unsupervised domain adaptation methods have been proposed. Liu \etal~propose to adapt the model to target domain with the guidance of outliers by collaborative learning \cite{liu2021generalizing}. Wang \etal~utilize contrastive learning to pull features with close gaze labels together \cite{wang2022contrastive}. Bao \etal~propose to improve the cross-dataset accuracy by the rotation consistency of gaze \cite{bao2022generalizing}. Nevertheless, the above methods require target domain images to train domain specific models, which is infeasible in real world settings, as target domain data is often inaccessible. Recently, Cheng \etal~propose to improve the generalization ability of gaze estimation model by purifying gaze feature in source domain\cite{cheng2022puregaze}. Without target domain images, the cross-domain gaze estimation problem is more challenging and remains much space to be explored.

%-------------------------------------------------------------------------

%------------------------------------------------------------------------

\begin{figure}
\begin{center}
% \begin{overpic} 
% [width=\linewidth]
% {example-image-a}
% \end{overpic}
\includegraphics[width=\linewidth]{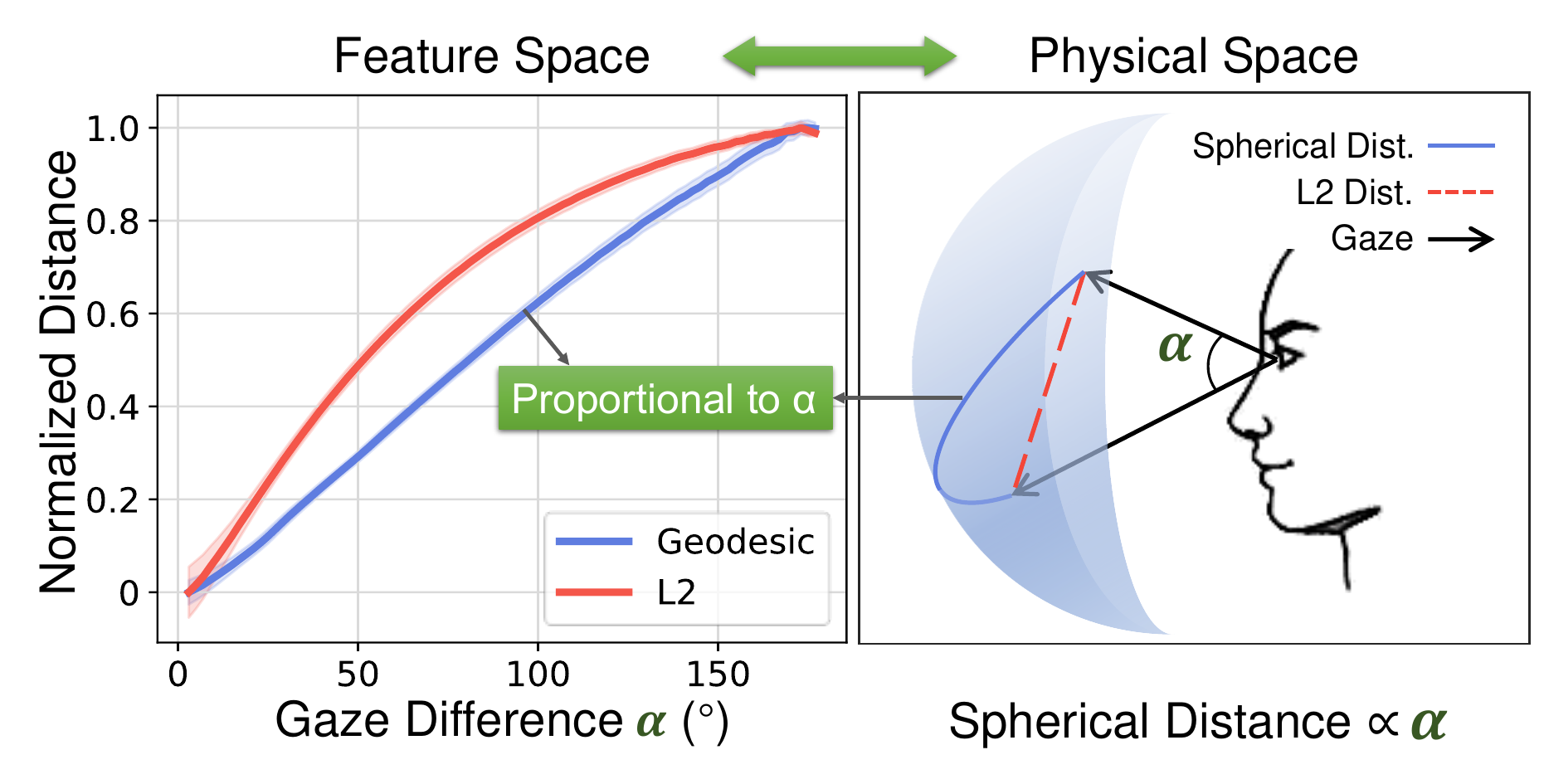}
\end{center}
% \vspace{-2mm}
\caption{
Left: the L2 and geodesic distance with respect to the angular difference between samples in ETH-XGaze dataset. Right: the physics of gaze. The geodesic distance between features and the distance along the spherical surface are proportional to the gaze differences.
}
\label{fig:Distance}
\vspace{2mm}
\end{figure}

\section{From Physical Rule to Regression Feature} \label{sec:three}
In this section, we introduce a concept of \feature~(\ft). Our motivation stems from a straightforward idea, i.e., for a \textit{regression task}, the features extracted by the network may exist in a way that they share the similar numerical patterns with the corresponding physical variables to be regressed. Once these features are discovered, they can be constrained by the same physical rules that lie in original variables. As the physical rules always hold true, this reduces the dependence on ground truth data in conventional methods, and thus helps enhance the interpretability of the algorithm, address the insufficiency of training data, overcome the overfitting problem of the algorithm, and improve the generalization ability.

\begin{figure}
\begin{center}
% \begin{overpic} 
% [width=\linewidth]
% {example-image-a}
% \end{overpic}
\includegraphics[width=\linewidth]{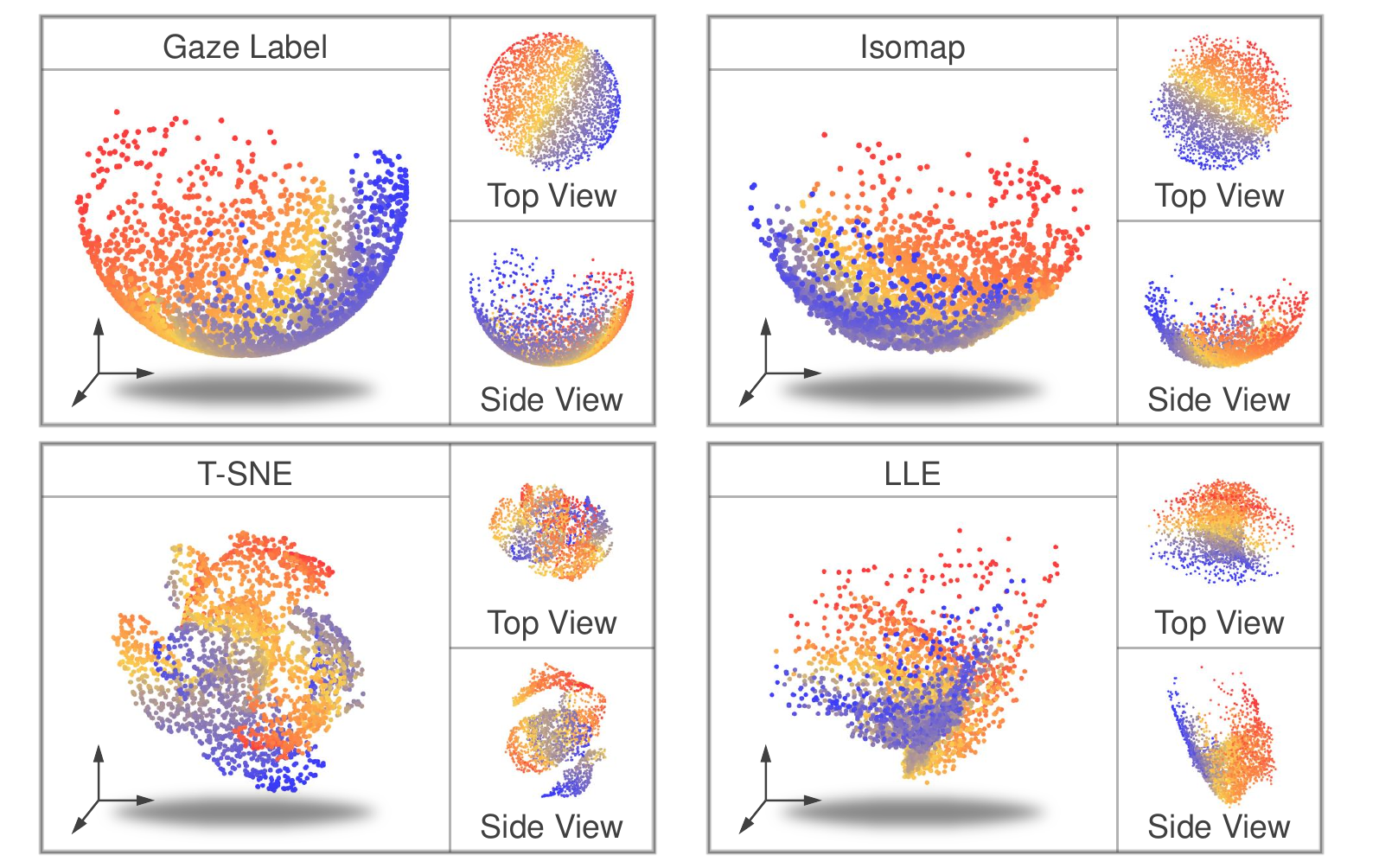}
\end{center}
% \vspace{-2mm}
\caption{
Distribution of gaze feature after dimension reduced to 3D space in ETH-XGaze dataset. Three different unsupervised manifold learning methods (Isomap, T-SNE and LLE) are used. Only the results of Isomap show identical patterns as the gaze ground truth.
}
\label{fig:Dimension}
\vspace{2mm}
\end{figure}

Motivated by this idea, this section explores the kind of features that should be in line with the laws of eye physiology for the gaze estimation problem. The gaze direction is commonly expressed as a 2D unit vector, so its angular rotation can be modeled as a displacement on a 2D unit sphere embedded in 3D Euclidean space. Thus, although the input to gaze estimation network is image data with million dimensions, its physics-consistent features mainly characterize 2D spherical motion. Consequently, we make the following \textit{Hypothesis:}

\begin{hyp}
The gaze features extracted from the images can be effectively mapped to the 2D unit spherical surface, and subject to the following physical consistency: the angle between the features (spherical distance) is proportional to the angle between the real-world gaze directions.
\end{hyp}

The gaze feature above is the feature extracted by the last convolutional layer of a ResNet-18 network trained on gaze estimation dataset, \ie~ETH-XGaze. Further details will be introduced in \cref{sec:method}.
In the following, we verify this hypothesis and give the computation of such physics-consistent 2D feature.

\begin{figure}
\begin{center}
% \begin{overpic} 
% [width=\linewidth]
% {example-image-a}
% \end{overpic}
\includegraphics[width=\linewidth]{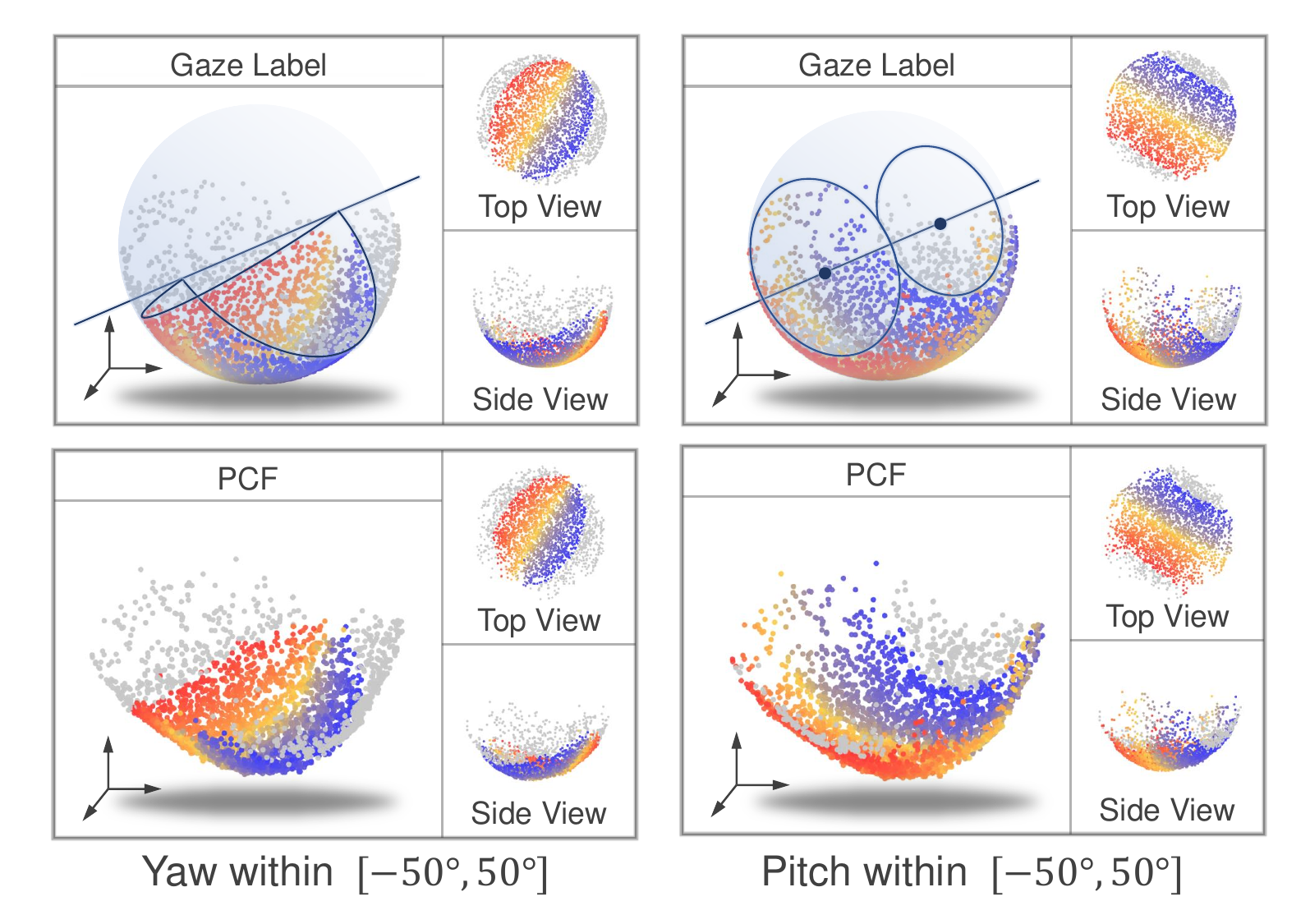}
\end{center}
% \vspace{-2mm}
\caption{
Gaze features after dimension reduction (\ie, PCF) share the same spherical distribution as gaze label. To better demonstrate the similarity, data points within certain gaze range are colored according to the gaze angles. In the visualization of PCF, the gaze pitch and yaw distribute along the longitudinal and latitudinal direction, just like the gaze labels.
% The results of Isomap colored according to the gaze angles of pitch and yaw respectively. 
}
\label{fig:IsoColored}
\vspace{2mm}
\end{figure}
\begin{figure*}
\begin{center}
% \begin{overpic} 
% [width=\linewidth]
% {example-image-a}
% \end{overpic}
\includegraphics[width=2\columnwidth]{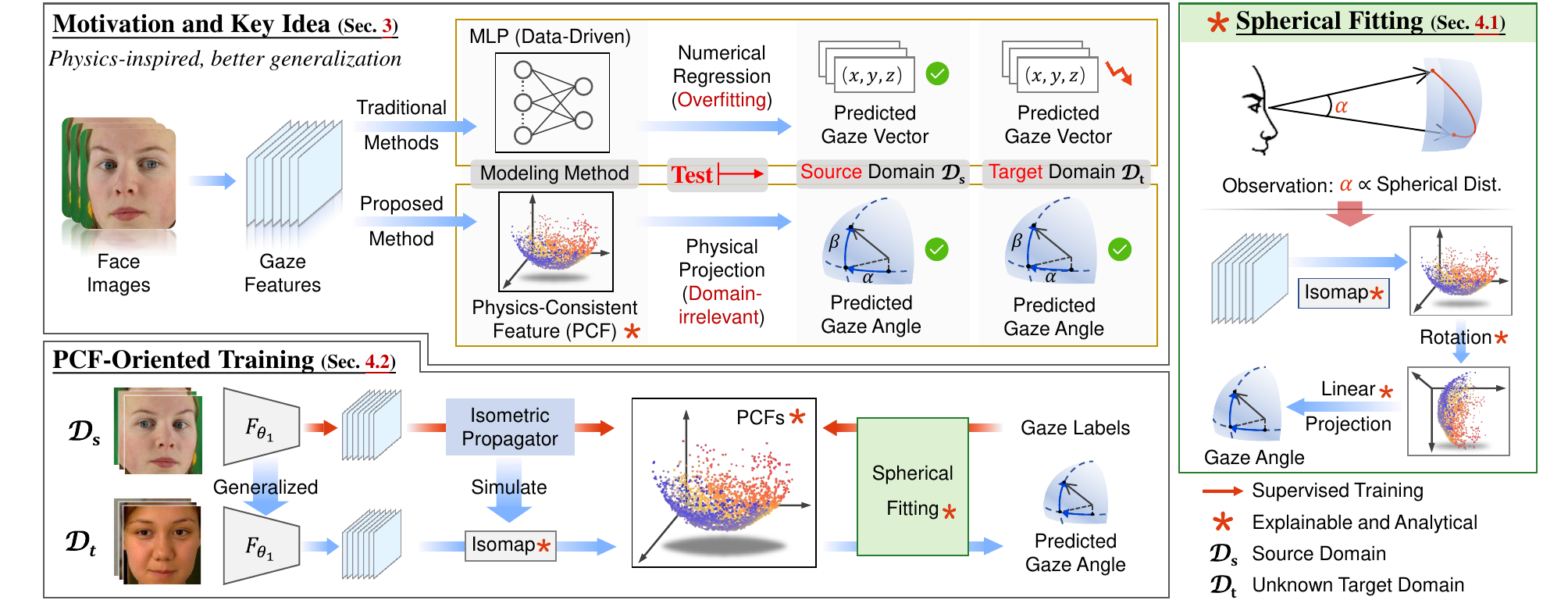}
\end{center}
% \vspace{-2mm}
\caption{
Overview of the proposed \prp~framework. The key idea of the \prp~framework is to improve the generalization ability by optimizing gaze model through physics of gaze, which is domain irrelevant. Based on the observation that the spherical distance is proportional to the gaze differences, we construct the \feature~(\ft) that connects gaze feature to the physics of gaze and estimates gaze by \proposeda~analytically. With the help of \ft~and \proposeda, the \prp~framework optimize gaze feature with the physics of gaze by the proposed \proposedb~for better generlization ability.
}
\label{fig:Method}
% \vspace{-2mm}
\vspace{2mm}
\end{figure*}
\subsection{Verification of the Hypothesis}
We conducted two prove-of-concept experiments to verify the above hypothesis. 

First, if the hypothesis holds, the Euclidean distance between features is proportional to the gaze angular difference only when gazes are close enough, \ie,
\begin{equation} \label{eq:eucdis}
    \Vert \bm{f}_i, \bm{f}_j \Vert_2 \propto \text{Ang}(\bm{y}^s_i,\bm{y}^s_j),
\end{equation}
where $\text{Ang}(\bm{y}^s_i,\bm{y}^s_j)$ is the angular difference between gaze ground truth $\bm{y}^s_i$ and $\bm{y}^s_j$. For samples with large gaze differences, spherical distances between features should be measured by Geodesic distance:
\begin{equation} \label{eq:geodesic}
    d_G(\bm{f}_i, \bm{f}_j) \propto \text{Ang}(\bm{y}^s_i,\bm{y}^s_j),
\end{equation}
where $d_G(\bm{f}_i, \bm{f}_j)$ is the geodesic distance between features.
A geodesic distance measures the shortest path between two features, and is computed by adding up small Euclidean distances (\cref{eq:eucdis}) of neighboring features along that path \cite{tenenbaum2000global}. Relationship in \cref{eq:geodesic} holds true globally in our hypothesis.
% \begin{equation}
%     Dis(f_i, f_j) \propto Ang(y_i, y_j)
% \end{equation}
% where $Dis(\cdot)$ is the target distance measurement, $Ang(y_i, y_j)$ is the angular difference between the ground truth gaze of sample $i, j$. 
As shown in the left part of \cref{fig:Distance}, the geodesic distance in feature space is linear to gaze differences globally in ETH-XGaze dataset. This numerical relationship of feature space is consistent with physical space, where the spherical distances is linear to gaze differences. The results verify our hypothesis that gaze feature constitutes a manifold in feature space, and is connected to gaze by geodesic distance.

Second, we map gaze features to 3D space by three different manifold learning methods. Specifically, we use the Isometric map (Isomap) \cite{isomap}, the Local Linear Embedding (LLE) \cite{lle}~and the T-distributed Stochastic Neighbor Embedding (t-SNE) \cite{tsne}. The results are shown in \cref{fig:Dimension}. Data points are colored according to the gaze rotation angles. For a more intuitive comparison, we also visualize the distribution of gaze ground truth in the first column. The results of Isomap distribute on a spherical surface, similar to the gaze ground truth. The results are consistent with our hypothesis as the Isomap maps features to low dimensional space according to the geodesic distance between them. To further investigate the distribution pattern, we only color data points within certain gaze range in \cref{fig:IsoColored}. It is clear that gaze pitch and yaw distribute along the longitudinal and latitudinal direction of the sphere, which is consistent with the physical definition of gaze. We name the 3D data points of Isomap as \feature~(\ft).
\ft~links gaze features to the physical definition of gaze by an intrinsic property of the gaze feature manifold, \ie, the geodesic distance. This insight of gaze features may facilitate the training of gaze estimation models, which will be discussed in next section.

\subsection{Benefit of \ft} \label{sec:benefit}
In this section, we discuss the application of \ft~and explain how it differs from common deep learning based gaze estimation approaches. In deep learning based approaches, gaze features are mapped to gaze directions using an Multilayer Perceptron (MLP). The MLP is trained based on enormous amount of data with ground truth. Thus, it suffers from the overfitting problem. 
%On the contrary, the proposed \ft~presents an explainable relation between gaze features and the gaze rotations.
On the contrary, the proposed \ft~connects features to gaze vectors by the intrinsic property of gaze feature manifold in an unsupervised manner. For a given sample, \ft~reveals where the ideal gaze feature should be in the feature space, so that it serves as a constraint to optimize gaze features in a more direct and fundamental way.

\section{The \proposed~Framework}
\label{sec:method}
The overview of the \prp~framework is shown in \cref{fig:Method}. Based on the analysis in \cref{sec:three}, we propose \proposeda, which estimates gaze from \ft~analytically by physical projection.
%First, We pretrain a normal gaze estimation model by L1 loss. We acquire \ft~by performing Isomap to source domain features and estimate gaze angles from \ft~by \proposeda. Then, we design an Iso Propagator to integrate the Isomap process into training scheme and retrain the gaze estimation model based on the guidance of \ft. The whole process is done in source domain without requiring any extra data.
With the help of \ft, we propose the \proposedb, which optimize the gaze feature space with the physic of gaze. As the physical rules are universal, the \proposedb~improves the generalization ability of gaze estimation model without extra training data.
\subsection{\proposeda}
As shown in \cref{fig:IsoColored}, \ft~of the pretrained model shares the same gaze distribution pattern as the ground truth. Thus, we propose \proposeda~(\pa) algorithm that estimates gaze angles according to the position of \ft~analytically. Specifically, we project \ft~to a sphere and fit the Euler angles of the projected \ft~to gaze angles by linear functions.

Given the source domain $\mathcal{D}_s = \{\bm{x}^s_i,\bm{y}^s_i|^{N_s}_{i=1}\}$ where $\bm{x}^s_i$ is the face image and $\bm{y}^s_i=(x_i, y_i, z_i)$ is the ground truth gaze direction vector, we pretrain the gaze estimation model by $\mathcal{L}_1$ loss function:
\begin{equation}
    \underset{\theta_1, \theta_2}{\arg \min} (\mathcal{L}_1( \bm{y}^s_i - L_{\theta_2}(F_{\theta_1}(\bm{x}^s_i)))|^{N_s}_{i=1}),
\end{equation}
where $F_{\theta_1}(\cdot)$ is the feature extractor CNN and $L_{\theta_2}(\cdot)$ is the regression MLP.  After the pretrain is completed, \ft~$\bm{f}_{pc}$ is constructed by performing Isomap on gaze features:
\begin{equation}
    \begin{aligned}
    \{\bm{f}_{pc,i}|^{N_s}_{i=1}\}& = \text{Isomap}(\{\bm{f}_i|^{N_s}_{i=1}\}),\\
    \bm{f}_i& = F_{\theta_1}(\bm{x}_i^s).
    \end{aligned}
\end{equation}
The \proposeda~algorithm estimates gaze from the \ft: $\bm{g}_i^s = \text{\pa}_{\theta_s}(\bm{f}_{pc,i})$.
To do so, we project $\bm{f}_{pc}$ to a sphere with center $\bm{O}_c$ and rotate it so that the orientation of the sphere could be aligned with gaze:
\begin{equation}
    (\bm{f}_{pc,i}')^T = \bm{R}(\bm{f}_{pc,i}-\bm{O_c})^T = (x'_i, y'_i, z'_i)^T,
\end{equation}
where $\bm{R}$ is the rotation matrix. The gaze estimation $\bm{g}_i^s=(\alpha_i, \beta_i)$ is the linear projections of the Euler angles of $\bm{f}'_{pc}$:
%Then, the pitch and yaw angle of $\bm{r'}_{pc,i}$ is mapped to gaze estimation $\bm{g}_i^s=(\alpha_i, \beta_i)$ with linear functions:
\begin{equation}
    \begin{aligned}
    &\alpha_i = k_1\arctan(\frac{-x'_i}{-z'_i})+b_1,\\
    &\beta_i = k_2\arcsin(-y'_i)+b_2.
    \end{aligned}
\end{equation}
Following the insight that geodesic distance between features is proportional to the gaze differences between samples, we estimate gaze directions from the gaze feature analytically by the proposed \proposeda~algorithm. The \proposeda~algorithm consists of 10 parameters $\theta_s=(\bm{O}_c, \bm{R}, k1, k2, b1, b2)$. These parameters are determined by optimizing the angular error between gaze estimation  and ground truth $\bm{y}^s$:
\begin{equation}
    \underset{\theta_s}{\arg \min} (\text{Ang}(\bm{g}_i^s, \bm{y_i^s})|^{N_s}_{i=1}).
\end{equation}
In practice, we randomly choose a subset of 2000 samples from the source domain for parameter optimization considering the time demand of Isomap. During inference, features of the new samples are concatenated to the geodesic distance map built in the training phase, which could be done in real time. \proposeda~algorithm generalizes well to target domains because the physics of gaze hold true globally.

\begin{table*}
\caption{
Cross-domain gaze estimation error in degrees. The proposed \proposed~method achieves improvements as large as $29.23\%$ without any extra training data. Percentage with underline indicates the largest improvement over corresponding baseline. Bold number indicates the lowest estimation error. $^*$ means that the number of neighbors in geodesic distance is customized for each source domain (100, 800 and 300 for \ETH, \GazeC~and \MPII~respectively). In row 5, the number is set to 300 for all source domain as default.
} % \caption
\vspace{2mm}
\renewcommand\arraystretch{1.1}
\centering
\resizebox{\linewidth}{!}{ %< auto-adjusts font size to fill line
\begin{tabular}{@{}cl|cc|cc|ccc@{}}

\toprule[1.3pt]
&Method &    \ETH $\rightarrow$\MPII  &  \ETH $\rightarrow$\ED   &  \GazeC$\rightarrow$\MPII   &  \GazeC$\rightarrow$\ED   &  \MPII $\rightarrow$\ETH &  \MPII $\rightarrow$\GazeC   &  \MPII $\rightarrow$\ED  \\
% \hline
\midrule[0.6pt]
1&Baseline \cite{cheng2022puregaze} &  8.13  &  7.74  &  9.89  &   11.42 & -   & -   & -   \\

2&PureGaze \cite{cheng2022puregaze}    &   7.08 \upscore{12.92}  &  7.48 \upscore{3.36}   &  9.28 \upscore{6.17}  &   9.32 \upscore{18.39}  &  -  &  -  &  -   \\
\midrule[0.6pt] \midrule[0.6pt]
% \hline \hline
3&Baseline    &  8.64  &  7.83   &  8.68  &  12.35  &  14.03  &  15.48  &  16.15  \\ 

4&Baseline+\pa  &  7.92 &  8.22   &  8.33  &  11.40  &  14.14  &  14.83  &  11.80  \\ %neighbor:300

5&\prp   &  7.40 \upscore{14.35} &   7.30 \upscore{6.77} &   8.36 \upscore{3.69}  &  10.18 \upscore{17.57}  &   13.69 \upscore{2.42}  &   13.76 \upscore{11.11}  &   12.40 \upscore{23.22}  \\ %neighbor:300
% \hline
\midrule[0.6pt]

%   & \multicolumn{2}{c|}{\prp~with 100 neighbors} & \multicolumn{2}{c|}{\prp~with 800 neighbors} & \multicolumn{3}{c|}{\prp~with 300 neighbors}\\
% & \multicolumn{7}{l|}{\prp~with 100, 800, 300 neighbors for \ETH, \GazeC and \MPII~respectively}\\
6&Baseline+\pa$^*$  &  7.98 &  8.45   &  7.79  &  10.97  &  14.14  &  14.83  &  11.80  \\ %neighbor:300

7&\prp$^*$   &  \marka{7.00} \upscorered{18.98} &  \marka{7.02} \upscorered{10.34} &  \marka{7.71} \upscorered{11.18}  &  \marka{8.74} \upscorered{29.23}  &  \marka{13.69} \upscorered{2.42}  &  \marka{13.76} \upscorered{11.11}  &  \marka{12.40} \upscorered{23.22}  \\

\bottomrule[1.3pt]
\end{tabular}
} %< \resizebox

\label{tab:BaseAccuracy}
\end{table*}

% baseline+\prp &  22.68 (0.31)  &  7.4 (1.24) &  7.3 (0.53)  &  20.87 (-1.73) &  8.36 (0.32)  &  10.18 (2.17)  &  13.69 (0.34)  &  13.76 (1.72)  &  11.67 (4.48)  \\ %neighbor:300
% PureGaze \cite{cheng2022puregaze}  & -   &  7.08 (1.05)  &  7.48 (0.26)  & -  &  9.28 (0.61)  &  9.32 (2.1)  &  -  &  -  &  -   \\
% Baseline+\pa~800 &  23.62  &  7.19  &  7.32  & 19.3  &  7.76  &  10.97  & 14.15   &  14.37  &  11.98  \\
% baseline+\prp~800 &  22.91  &  7.36  &  7.4  &  21.16 &  7.71  &  8.74  &  12.66  &  14.46  &   13.44 \\
% Baseline+\pa~100 &  25.88  &  7.98  &  8.45  & 18.67  &  8.28  &  13.34  &  14.04  &  14.83  &  12.99  \\
% baseline+\prp~100 &  22.76  &   7.00 &  7.02  &  19.33 &  8.63  &  15.28  &  14.74  &  14.49  &  12.02  \\
\begin{table}
\caption{
Cross domain gaze estimation error in degrees. $\dag$ indicates that the method employs ResNet-50 as backbone. The proposed \prp~improves accuracy of different baseline model and outperforms other SOTA methods.
} % \caption
\vspace{2mm}
\renewcommand\arraystretch{1.2}
\centering
\resizebox{\linewidth}{!}{ %< auto-adjusts font size to fill line
\begin{tabular}{@{}l|cccc@{}}

\toprule[1.3pt]
Method   &  \ETH $\rightarrow$\MPII  &  \ETH $\rightarrow$\ED  &  \GazeC $\rightarrow$\MPII   &  \GazeC $\rightarrow$\ED  \\
\midrule[0.8pt] 
Full-Face\cite{zhang2017s} & 12.35 & 30.15 & 11.13 & 14.42 \\
ADL\cite{kellnhofer2019gaze360} & 7.23 & 8.02 & 11.36 & 11.86 \\
CA-Net\cite{cheng2020coarse} & - & - & 27.13 & 31.41 \\
LatentGaze\cite{lee2022latentgaze} & 7.98 & 9.81 & - & - \\
PureGaze\cite{cheng2022puregaze} & 7.08 & 7.48 & 9.28 & \marka{9.32} \\
\midrule[0.8pt] 
Baseline &  8.64  &  7.83    &  8.68  &  12.35  \\ 
\prp & 7.40 & \marka{7.30} & 8.36 & 10.18\\
\midrule[0.8pt] 
Baseline$\dag$ &  7.3  &  8.43    &  8.22  &  11.79  \\ 
% % GVBGD \cite{cui2020gradually}&  7.64  &  12.44    &  6.68  &  7.27\\
% PnP-GA \cite{liu2021generalizing} &  5.53 & 5.87 & 6.18  & 7.92 \\
% CRGA \cite{wang2022contrastive} &  5.68 & \marka{5.72} & 6.09  & 6.68 \\
% RUDA \cite{bao2022generalizing} &  5.70 & 6.29 & 6.20  & \marka{5.86} \\
% \midrule[0.8pt] 

\prp$\dag$ &  \marka{6.15}  &  7.66    &  \marka{7.61}  &  11.42  \\ 
\bottomrule[1.3pt]
\end{tabular}
} %< \resizebox

\label{tab:comparison2SOTA}

\end{table}

\subsection{\proposedb}
Our ultimate goal is to train the feature extractor $F_{\theta_1}$ under the guidance of the \proposeda~algorithm . Unfortunately, Isomap is hard to integrate to the back propagation process because it is both time and space consuming. The time complexity of Isomap is $N^2log_N$ and the space demand is $N^2$, where $N$ is the number of samples. The time and memory space required by the Isomap to process hundreds of thousands of gaze features is enormous. Thus, we propose the \proposedb~(\pb) method to solve this problem. 

In \proposedb~phase, we train a Isometric Propagator to replace the Isomap algorithm. First, we perform Isomap on a subset of source domain with $N'$ samples:
\begin{equation}
    \{\bm{f}_{pc,i}|^{N'_s}_{i=1}\} = \text{Isomap}(\{\bm{f}_i|^{N'_s}_{i=1}\}).
\end{equation}
$N'$ is set to $2000$ as it is enough to cover the common gaze range. Then, we train a simple three layer MLP as the Isometric Propagator $\text{IP}_{\theta_3}(\cdot)$ to replace the Isomap algorithm. $\text{IP}_{\theta_3}(\cdot)$ is supervised by $\mathcal{L}_1$ loss function:
\begin{equation}
    \underset{\theta_3}{\arg \min} (\mathcal{L}_1(\bm{f}_{pc,i}, \text{IP}_{\theta_3}(\bm{f}_i))|^{N'_s}_{i=1}).
\end{equation}
The idea is to simulate the Isomap algorithm by training the Isometric Propagator. During \proposedb, \ft~is derived by the Isometric Propagator instead of Isomap algorithm: $\bm{f}_{pc,i}=IP_{\theta_3}(\bm{f}_i))$. After the training of Isometric Propagator, we freeze the parameters of it and optimize the model by constraining \ft. As the \proposeda~algorithm is analytical, we inversely calculate the ground truth \ft~$\hat{\bm{f}}_{pc,i}$ from the gaze ground truth:
\begin{equation}
    \hat{\bm{f}}_{pc,i} = \text{\pa}_{\theta_s}^{-1}(\bm{y}_i).
\end{equation}
Finally, we retrain the feature extractor by constraining the \ft~with $\mathcal{L}_1$ loss function:
\begin{equation}
    \underset{\theta_1}{\arg \min} (\mathcal{L}_1(\hat{\bm{f}}_{pc,i}, \text{IP}_{\theta_3}(F_{\theta_2}(\bm{x}_{i}))|^{N_s}_{i=1})).
\end{equation}
In this way, we directly optimize gaze features according to their relation to physical definition of gaze. As the optimization objective of the gaze features is to match the distribution modeled by \ft~and \proposeda, user gaze is estimated by Isomap and $\text{\pa}_{\theta_s}$ at inference time:
\begin{equation}
    \bm{g}_i = \text{\pa}_{\theta_s}(\text{Isomap}(F_{\theta_1}(\bm{x}_i))).
\end{equation}
%%%%%%%%% REFERENCES

\subsection{Implementation Details}
We employ PyTorch for implementation. ResNet 18 is used as the baseline following previous studies \cite{cheng2022puregaze, wang2022contrastive, bao2022generalizing}. For the training of the pretrain model, IP and \proposedb, we use the Adam optimizer with a learning rate of $10^{-4}$. The model is pretrained for 10 epochs. The \proposedb~is also 10 epochs, while the IP is trained for 100 epochs. For Isomap, we use the implementation of Scikit-learn. The number of neighbor in geodesic distance is set to 300 by default.

\section{Experiments}
\subsection{Data Preparation}
We conduct experiments on four commonly used gaze estimation datasets: ETH-XGaze (\ETH) \cite{zhang2020eth}, Gaze360 (\GazeC) \cite{kellnhofer2019gaze360}, MPIIFaceGaze (\MPII) \cite{zhang2017s} and EyeDiap (\ED) \cite{funes2014eyediap}. We pre-process the data following the techniques in \cite{liu2021generalizing, cheng2021appearance}.

\textbf{ETH-XGaze:} $756k$ images captured by high resolution cameras in laboratory environment with large gaze range.

\textbf{Gaze360:} $101k$ images captured by a $360\degree$ camera on streets with large gaze range.

\textbf{MPIIFaceGaze:} $45k$ images captured by web camera during daily usage of laptop computers. The gaze range of \MPII~is less than half the range of \ETH~and \GazeC. Thus, in \MPII$\rightarrow$\ETH~and \MPII$\rightarrow$\GazeC~ settings, we only use samples within the gaze range of \MPII~for testing.

\textbf{EyeDiap:} $16k$ images captured under laboratory environment with screen and floating targets. As the number of images is significantly less than other datasets, we only use \ED~as target domain.

In addition, the cross-domain error between \ETH~and \GazeC~is extremely large (around $20\degree$). Thus, we exclude the \ETH$\rightarrow$\GazeC~and \GazeC$\rightarrow$\ETH~settings in our experiments, which is also excluded in previous studies \cite{wang2019generalizing, liu2021generalizing, bao2022generalizing, cheng2022puregaze}.

\subsection{Performance of \prp~Framework} \label{sec:performance}
\subsubsection{Domain Generalization Accuracy}\label{sec:accuracy}
To test the generalization ability of the \prp~framework, we conduct experiments in 6 different cross domain settings. The results are shown in \cref{tab:BaseAccuracy}. To make a fair comparison with \cite{cheng2022puregaze}, we provide the error of baseline reported in the original paper and the percentage of improvements. The \prp~framework in the default setting achieves stable improvements in all 7 settings, proves the effectiveness of the \prp~framework. Compared to the baseline model, the proposed framework achieves $23.22\%$ improvement in \MPII$\rightarrow$\ED~setting. 
Row 4 of \cref{tab:BaseAccuracy}~shows that without \proposedb, the proposed \proposeda~still improves the cross domain performance in most settings (5 out of 7). These results prove  our point in \cref{sec:benefit} that by estimating gaze analytically through the physics of gaze, \proposeda~is less affected by overfitting than the regression MLP. The performance of \prp~is further improved by tuning the number of neighbors in geodesic distance for each source domain. The numbers of neighbors are set to 100, 800 and 300 for \ETH, \GazeC~and \MPII~respectively. As shown in the row 7 of \cref{tab:BaseAccuracy}, the improvements \prp$^*$ brought are over $50\%$ more than PureGaze. 
%Overall, cross domain experiments prove that the proposed \proposeda~produces more accurate estimation than regression MLP in cross domain settings. By optimizing the model with \ft and ~\proposedb, the generalization ability of the baseline model is further improved. 

\begin{table}
\caption{
The mean and standard deviation of the estimation error for last 5 epochs. The proposed \prp~are not only more accurate but also more stable than the baseline. The \proposeda~is also more stable than the regression MLP by estimating gaze analytically based on physics of gaze.
} % \caption
\vspace{2mm}
\renewcommand\arraystretch{1.2}
\centering
\resizebox{\linewidth}{!}{ %< auto-adjusts font size to fill line
\begin{tabular}{@{}l|cccc@{}}

\toprule[1.3pt]
Method   &  \ETH $\rightarrow$\MPII  &  \ETH $\rightarrow$\ED  &  \GazeC $\rightarrow$\MPII   &  \GazeC $\rightarrow$\ED  \\
\midrule[0.8pt] 
Baseline&  8.66\std{0.52}  &  7.76\std{0.83}    &  8.59\std{0.65}  &  10.87\std{1.40}  \\ 
% % GVBGD \cite{cui2020gradually}&  7.64  &  12.44    &  6.68  &  7.27\\
% PnP-GA \cite{liu2021generalizing} &  5.53 & 5.87 & 6.18  & 7.92 \\
% CRGA \cite{wang2022contrastive} &  5.68 & \marka{5.72} & 6.09  & 6.68 \\
% RUDA \cite{bao2022generalizing} &  5.70 & 6.29 & 6.20  & \marka{5.86} \\
% \midrule[0.8pt] 
Baseline+\pa  &  7.88\std{0.21}  & 7.72\std{0.30}  &  8.57\std{0.39} & 10.94\std{\underline{0.80}} \\ 
\prp & \marka{7.44}\std{\underline{0.16}} & \marka{7.3}\std{\underline{0.12}} & \marka{8.33}\std{\underline{0.29}} &\marka{9.35}\std{0.82}\\
\bottomrule[1.3pt]
\end{tabular}
} %< \resizebox

\label{tab:Stability}
\end{table}
\begin{table}
\caption{
Additional experiments: fine-tuning models with 100 target domain images for domain specific generalization
} % \caption
\vspace{2mm}
\renewcommand\arraystretch{1.1}
\centering
\resizebox{\linewidth}{!}{ %< auto-adjusts font size to fill line
\begin{tabular}{@{}l|cccc@{}}

\toprule[1.3pt]
Method   &  \ETH $\rightarrow$\MPII  &  \ETH $\rightarrow$\ED  &  \GazeC $\rightarrow$\MPII   &  \GazeC $\rightarrow$\ED  \\
\midrule[0.8pt] 
Baseline&  8.64  &  7.83    &  8.68  &  12.35  \\ 
% % GVBGD \cite{cui2020gradually}&  7.64  &  12.44    &  6.68  &  7.27\\
% PnP-GA \cite{liu2021generalizing} &  5.53 & 5.87 & 6.18  & 7.92 \\
% CRGA \cite{wang2022contrastive} &  5.68 & \marka{5.72} & 6.09  & 6.68 \\
% RUDA \cite{bao2022generalizing} &  5.70 & 6.29 & 6.20  & \marka{5.86} \\
% \midrule[0.8pt] 
Fine-tune  &  5.11  &  5.92  &  5.59 & \marka{ 6.30} \\ 
\prp-Ft & \marka{4.74} & \marka{5.88} &\marka{5.41} & 6.39\\
\bottomrule[1.3pt]
\end{tabular}
} %< \resizebox

\label{tab:Adaptation}

\end{table}

In \cref{tab:comparison2SOTA}, we compare the \prp~with SOTA gaze estimation methods \cite{zhang2017s,kellnhofer2019gaze360,cheng2020coarse}~and gaze generalization methods \cite{cheng2022puregaze,lee2022latentgaze}. For \cite{zhang2017s,kellnhofer2019gaze360,cheng2020coarse,cheng2022puregaze}, we report the cross-domain accuracy from \cite{cheng2022puregaze}. We report the accuracy of \cite{lee2022latentgaze} from the original paper. The results indicate that the proposed \prp~outperforms SOTA gaze estimation and gaze generalization methods. Results in the bottom rows also show that \prp~achieves stable improvement across different backbone models (ResNet-18 and ResNet-50).
Overall, above experiments prove that the \prp~effectively improves the cross-domain gaze estimation accuracy.

\subsubsection{Stability of the \prp}
%In practical settings, target domain data or label is often not available. There is no way to know which training epoch would end up with best accuracy on target domain. Thus, stability of the model is also important in cross-domain settings. If the accuracy of the model jitters severely, the training could stop at an epoch with large estimation error by accident. 
In practical settings, target domains are often unknown, makes the stability of the model critical. Stable performance across different epochs means less risk of randomly stopping at an epoch with large estimation error.
To verify the stability of the \prp, we calculate the mean and standard deviation (std.) of the gaze estimation error for different methods in \cref{tab:Stability}. We choose the last 5 epochs to make sure the model is converged. Results show that the std. of \prp~is less than half of the baseline in most settings (3 out of 4). Besides, although the accuracy of \proposeda~is similar to the regression MLP (baseline) without \proposedb, it still appear to be much more stable because \proposeda~estimates gaze analytically from physical rules instead of data-driven methods. Note we strictly follow the cross-domain condition and report the accuracy of the last epoch in \cref{tab:BaseAccuracy}~and \cref{tab:comparison2SOTA}.

\begin{figure}
\begin{center}
% \begin{overpic} 
% [width=\linewidth]
% {example-image-a}
% \end{overpic}
\includegraphics[width=\linewidth]{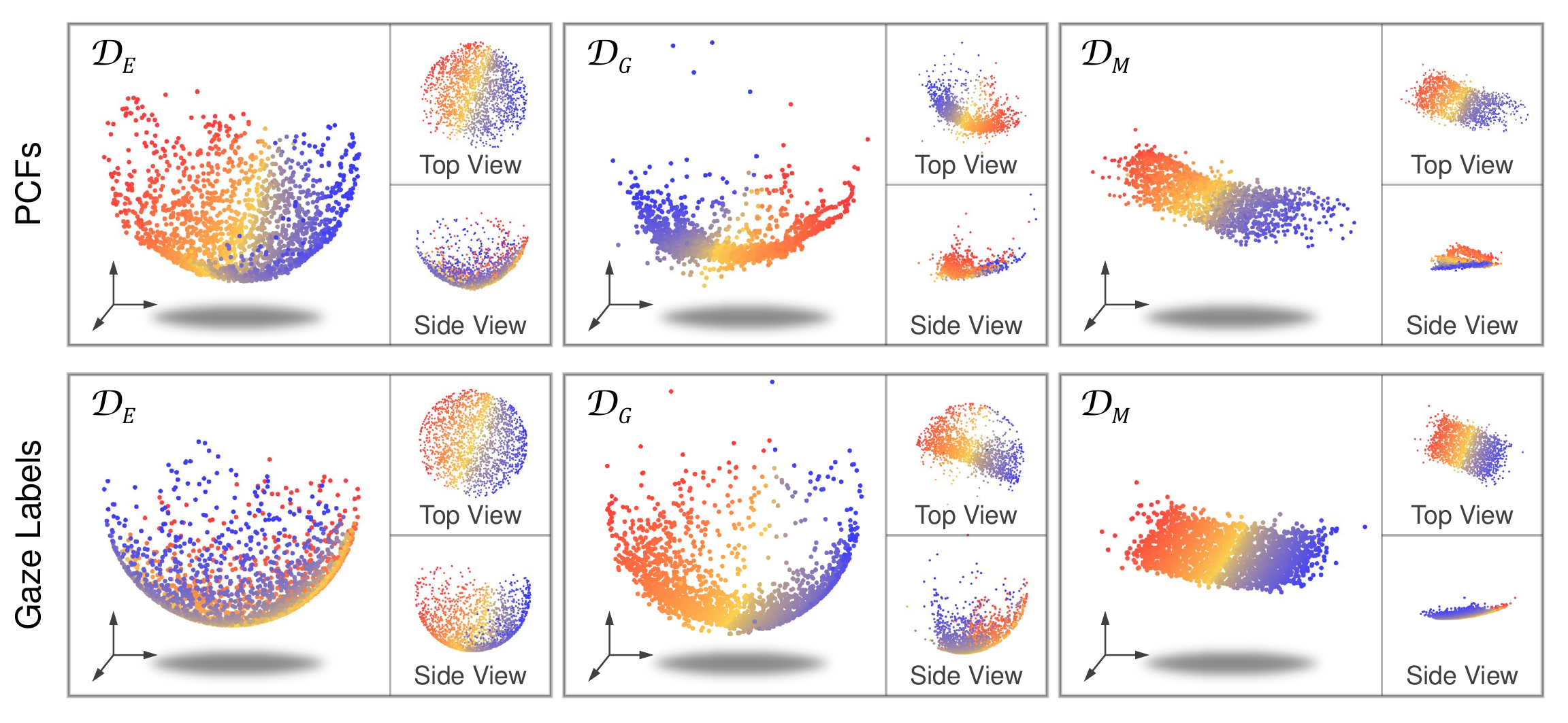}
\end{center}
% \vspace{-2mm}
\caption{
Visualization of the PCF and gaze label from \ETH, \GazeC~and \MPII. PCF from three datasets all share the same distribution pattern with gaze ground truth despite significant differences (image quality, head poses, gaze range, \etc) between datasets. It proves the generalization ability of \prp.
}
\label{fig:DatasetSpheres}
\vspace{2mm}
\end{figure}

\subsubsection{Additional Experiment: Domain Adaptation with Target Domain Data}
Results in \cref{sec:accuracy} show that the \prp~framework improves the generalization ability significantly without target domain data. To further prove the advantage of \prp~framework, we conduct additional domain adaptation experiments with 100 target domain images. We fine-tune the baseline model with a learning rate of $5*10^{-5}$ for 50 epochs in target domain. For \prp, we train the model following the~\proposedb for 30 epochs and calculate parameters of \proposeda~with the 100 target domain images, which is named \prp-Ft. The results are shown in \cref{tab:Adaptation}. The \prp-Ft outperforms fine-tune method in most settings. The results validate the advantage of \prp~over fine-tune method in supervised domain adaptation with small amount of samples (100), although \prp~is not designed for such purpose.

\begin{figure}
\begin{center}
% \begin{overpic} 
% [width=\linewidth]
% {example-image-a}
% \end{overpic}
\includegraphics[width=\linewidth]{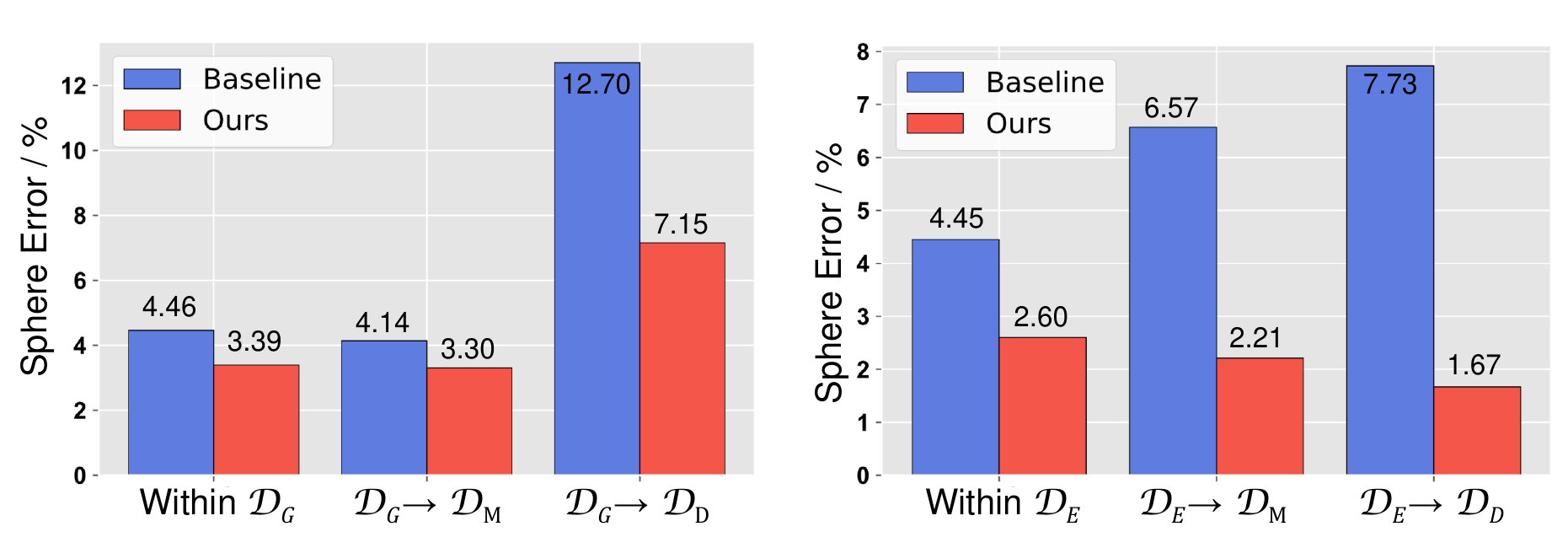}
\end{center}
% \vspace{-2mm}
\caption{
The Sphere Error of of the \pa~before (Baseline) and after \proposedb~(Ours). Smaller sphere error indicates that the \ft s are more consistent with the physics of gaze after \proposedb.
%\proposedb~significantly reduce the sphere error of PCF. The results indicate that after \proposedb, the \ft s are more consistent with the physics of gaze.
}
\label{fig:Sphere}
% \vspace{-2mm}
\vspace{2mm}
\end{figure}

\subsection{Verification of the \prp}
In \cref{sec:performance}, we show the effectiveness of the \prp~by cross domain gaze estimation accuracy. In this section, we verify the idea behind \prp. The \prp~works only if the following propositions hold true: 1) \ft~shares the same gaze distribution pattern as the gaze label; 2) \proposedb~constrains the model to extract more physics-consistent features. We verify these two key propositions statistically and intuitively in this section.

\textbf{Proposition (1).} \cref{fig:IsoColored} proves that \ft s in \ETH~share the same gaze distribution pattern as the gaze label, we further analyze \ft s in other within and cross dataset settings. We visualize \ft s of the baseline model in \ETH, \GazeC~and \MPII~under within dataset settings in \cref{fig:DatasetSpheres}. It is clear that the shape and range of \ft s is consistent with gaze labels. The gaze (visualized in different color) changes smoothly along the spherical surface of \ft s, proves that the \ft~is well-connected to gaze. \cref{fig:DatasetSpheres}~proves that the Hypothesis 1 holds true when the image quality and gaze range of the training dataset change dramatically.
For \ft s under cross dataset settings, results in row 4 and 6 of \cref{tab:BaseAccuracy} already show that \ft~is well mapped to gaze by \proposeda, \ie, rotation and linear fitting in all cross domain settings. Thus, proposition (1) holds true in both within and cross dataset settings.

\begin{figure}
\begin{center}
% \begin{overpic} 
% [width=\linewidth]
% {example-image-a}
% \end{overpic}
\includegraphics[width=\linewidth]{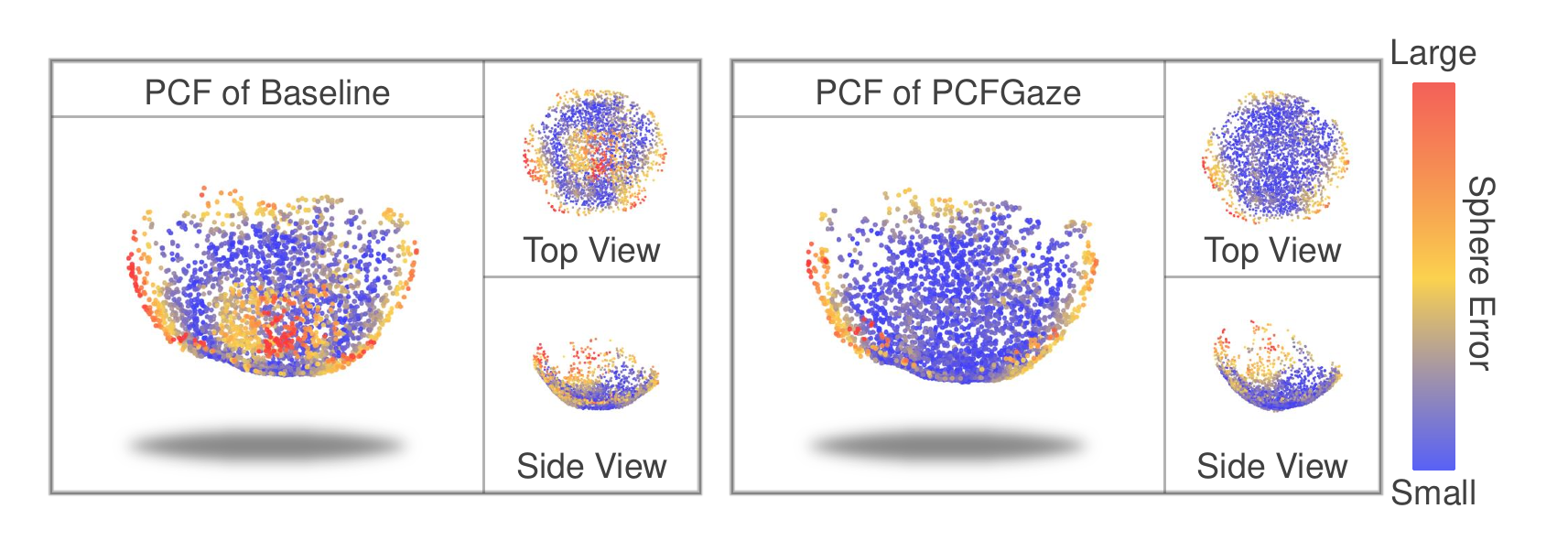}
\end{center}
% \vspace{-2mm}
\caption{
Visualization of the Sphere Error before and after \proposedb~in \ETH. The Sphere Error of \prp~is obviously smaller, proves that gaze feature extracted by \prp~is more consistent with gaze ground truth.
}
\label{fig:VisSphere}
% \vspace{-2mm}
\vspace{2mm}
\end{figure}

\textbf{Proposition (2).} We verify proposition (2) by analyzing whether the extracted features after \proposedb~are more consistent with the physics of gaze. As high dimensional features are hard to visualize and understand, we analyze \ft s for replacement. To see how consistent \ft s are with the physics of gaze, we measure how spherical \ft s are by calculating the Sphere Error, \ie, the average distance from \ft s to the spherical surface. As shown in \cref{fig:Sphere}, the Sphere Error of the \prp~is lower than the baseline in both within and cross domain settings. For more intuitive understanding, we visualize the Sphere Error of \ft s from \ETH~before and after \proposedb. The Sphere Error is significantly reduced in the center area after \proposedb. Besides, the side views also show that the surface of the \ft s after \proposedb is more spherical. Thus, we verify that proposition (2) holds true from both objective and subjective perspectives.

\subsection{Additional Application of \prp: Data Analysis}

\begin{figure}
\begin{center}
% \begin{overpic} 
% [width=\linewidth]
% {example-image-a}
% \end{overpic}
\includegraphics[width=\linewidth]{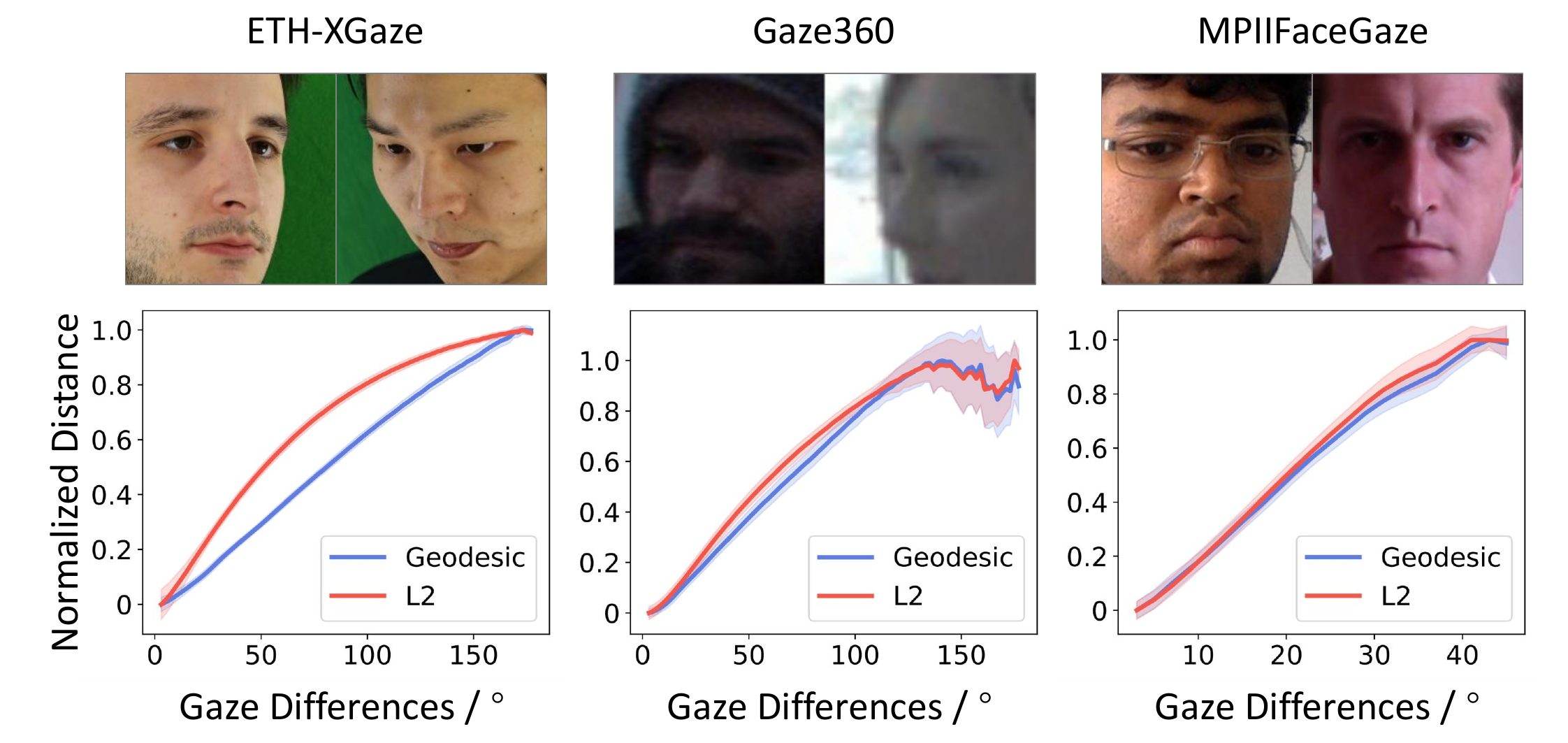}
\end{center}

\caption{
The L2 and Geodesic distances between gaze features with respect to the angular difference between samples from \ETH, \GazeC~and \MPII. 
}
\label{fig:DatasetDistances}
\vspace{2mm}
\end{figure}

As the \prp~reveals how gaze feature is connected to the physics of gaze in an analytical way, ideas behind \prp~could also serve as criteria during data analysis. In \cref{fig:DatasetDistances}, we visualize the $\mathcal{L}_2$ and geodesic distances with respect to the gaze difference between samples from \ETH, \GazeC~and \MPII. The results show the obvious differences between datasets. 
%\cref{eq:geodesic} holds true across all sample pairs in \ETH. The pattern is clear and the standard deviation (std) is small. We assume it is because the high resolution of the images and the controlled environment during collecting. For \GazeC, the pattern holds within $0\degree$ to around $120\degree$ range. When the gaze differences exceed $120\degree$, the pattern become random and the std increases dramatically. As the pattern holds in this range for \ETH, we assume the possible reason is that for \GazeC, the quality of samples in the edge of the gaze range decreases. So that extracted features from these samples do not behavior consistently with the others. For \MPII, the pattern is less obvious. One possible reason is that the gaze range of \MPII~is significantly smaller than \ETH~and \GazeC. The gaze distribution area of \MPII~can be seen as a small neighbor area compared to other two dataset. Overall, \cref{fig:DatasetDistances}~shows that the feature space of \ETH~is more consistent with the gaze physics. It coincides with the results that baseline model trained in \ETH~achieves better cross-domain performance than \GazeC~and \MPII.
\cref{eq:geodesic} holds true across all sample pairs in \ETH. We assume it is because the high resolution  and the controlled environment during collecting. For \GazeC, the pattern become random and the std increases dramatically when the gaze differences exceed $120\degree$. We assume the possible reason is that  the quality of samples in the edge of the gaze range decreases, makes the corresponding features more random. For \MPII, the pattern is less obvious. One possible reason is that the narrow gaze range of \MPII~could be seen as a small neighbor area compared to other two dataset. Overall, \cref{fig:DatasetDistances}~shows that the feature space of \ETH~is more consistent with the gaze physics. Correspondingly, the \ETH~baseline performs the best in cross-domain evaluation.

\section{Limitations and Future Work}
% Iso如何结合进训练流程
The \prp~successfully alleviates one of the major challenges in cross-domain settings, \ie, the absence of target domain data. Another challenge is that the gaze range of domains are different. The improvement of \prp~is less obvious if the gaze range of the target domain is significantly larger than the source domain. Existing methods still can not handle such problem well. Although theoretically the \proposeda~is capable of estimating unseen gaze range, the feature extractor can not handle samples with unseen gaze range correctly. In the future, we may dive deeper to the gaze feature space and try to tackle this problem.

\section{Conclusion}
In this paper, we propose the \prp~framework to increase the generalization ability of gaze estimation models without target domain data. We connect gaze features to the physics of gaze analytically through \ft, and constrains the gaze estimation model by the physics of gaze. As the physics of gaze is fundamental and explainable, the constrain of \prp~is less affected by the overfitting problem. Experimental results show that the \prp~framework improves accuracy in 8 different cross-domain settings. The idea of \ft~may inspire method design for other physical related regression tasks such as pose estimation.

{\small
\bibliographystyle{ieee_fullname}
\bibliography{egbib}
}

\end{document}